\pdfoutput=1

\documentclass[11pt]{article}

\usepackage[]{acl}

\usepackage{times}
\usepackage{latexsym}

\usepackage[T1]{fontenc}

\usepackage[utf8]{inputenc}
\usepackage{graphicx}
\usepackage{amsmath}
\usepackage{xcolor}
\usepackage{booktabs}
\usepackage{booktabs}
\usepackage{multirow}

\usepackage{microtype}

\title{Towards Socially Intelligent Agents with Mental State Transition and Human Value}

\author{Liang Qiu$^*${\normalfont\textsuperscript{1}}, Yizhou Zhao$^*${\normalfont\textsuperscript{1}}, Yuan Liang{\normalfont\textsuperscript{2}}, Pan Lu{\normalfont\textsuperscript{1}}, Weiyan Shi{\normalfont\textsuperscript{3}}, Zhou Yu{\normalfont\textsuperscript{3}}, Song-Chun Zhu{\normalfont\textsuperscript{1}} \\
  {\normalfont\textsuperscript{1}}UCLA Center for Vision, Cognition, Learning, and Autonomy \\    
  {\normalfont\textsuperscript{2}}University of California, Los Angeles \\
  {\normalfont\textsuperscript{3}}Columbia University \\
  \texttt{liangqiu@ucla.edu}
}

\begin{document}
\maketitle
\def\thefootnote{*}\footnotetext{Equal contribution. The work was done prior to Liang joining Amazon Alexa.}\def\thefootnote{\arabic{footnote}}
\begin{abstract}
Building a socially intelligent agent involves many challenges. One of which is to track the agent's mental state transition and teach the agent to make decisions guided by its value like a human. Towards this end, we propose to incorporate mental state simulation and value modeling into dialogue agents. First, we build a hybrid mental state parser that extracts information from both the dialogue and event observations and maintains a graphical representation of the agent's mind; Meanwhile, the transformer-based value model learns human preferences from the human value dataset, \textsc{ValueNet}. Empirical results show that the proposed model attains state-of-the-art performance on the dialogue/action/emotion prediction task in the fantasy text-adventure game dataset, LIGHT. We also show example cases to demonstrate: (\textit{i}) how the proposed mental state parser can assist the agent's decision by grounding on the context like locations and objects, and (\textit{ii}) how the value model can help the agent make decisions based on its personal priorities.
\end{abstract}

\section{Introduction}
Recently, there has been remarkable progress in language modeling with large-scale pretrained models~\citep{vaswani2017attention, devlin-etal-2019-bert, radford2019language}. Such models are used to build either general chatbots~\citep{zhang-etal-2020-dialogpt} or task-oriented dialogue systems~\citep{peng2020soloist, acharya2021alexa, qiu-etal-2020-structured}. While most of these systems have been able to generate fluent sentences, there are two major challenges towards building socially intelligent agents. 
\begin{figure}[ht]
\begin{center}
\centerline{\includegraphics[width=0.9\linewidth]{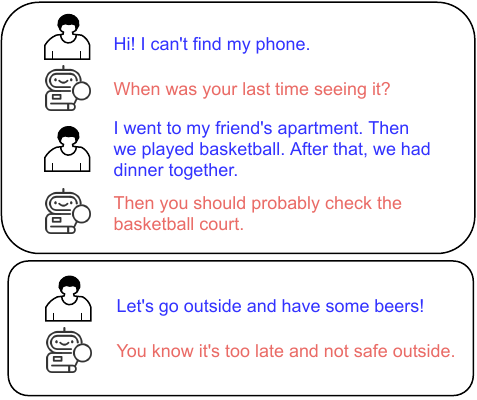}}
\caption{Socially intelligent agents with mental state simulation and human values.}
\label{fig:teaser}
\end{center}
\vspace{-10mm}
\end{figure}
First, considering dialogues as a "meeting of minds"~\citep{gardenfors2014geometry} or achieving some alignment of the interlocutors' mental models~\citep{rumelhart1986sequential, stolk2016conceptual}, few existing works are explicitly tracking the mental state transition of agents~\citep{adhikari2020learning}. Endowing current dialogue systems with such capability would allow the agent to condition its utterance on the context, simulate the effect of its actions, and further help understand the extended meaning, implicature, and irony expressed by the user~\citep{grice1981presupposition, grice1989indicative}. Second, it remains under-explored to teach agents to make a rational decision guided by its value. From a social and cultural perspective, humans tend to have a common preference described by the utility function related to individual values, common sense, and social awareness. For the example in Figure \ref{fig:teaser}, someone who values personal security prefers staying at home rather than going outside at night.

Our work aims to alleviate the aforementioned problems, based on Embodied Cognitive Linguistics (ECL)~\citep{lakoff1980metaphorical, gardenfors2014geometry} and established value theories in sociology~\citep{schwartz2012overview}. The ECL states that natural language is inherently executable, driven by mental simulation and metaphoric inference~\citep{lakoff1980metaphorical}, and learned through embodied interaction~\citep{feldman2004embodied, tamari-etal-2020-language}. Following its tenents, we present a hybrid mental state parser that converts dialogue and event observations into a graphical representation of the agents' mind. Initialized with the location and object description, the interpretable representation is updated through the interaction history to track the evolving process of an agent's belief about surroundings and other agents. 

In the field of intercultural research, \citet{schwartz1992universals, schwartz2012refining} identify basic individual values that are recognized across cultures. Inspired by the theory, we propose to incorporate a value model that learns social common preferences from the human value knowledge base, \textsc{ValueNet}~\cite{Qiu_Zhao_Li_Lu_Peng_Gao_Zhu_2022}. We perform experiments on a large-scale text-based embodied environment LIGHT~\citep{urbanek-etal-2019-learning}. Empirical results show that the model with our mental state emulator and value function achieves the highest performance that aligns with human annotation among existing transformer-based models. Moreover, case studies further demonstrate that the mental state provides extra context information, while the value model helps agents make value-driven decisions.

Our contributions are two-fold. First, we propose to rethink the design of current dialogue systems and suggest a new paradigm from the perspective of cognitive science and contemporary sociology. Second, we present a new framework for building socially intelligent agents by incorporating mental state simulation and human value modeling into dialogue generation and decision making. Our methodology can be generalized to a wide range of interactive social situations in dialogue systems~\citep{zhao2019socially}, virtual reality~\citep{lai2019social}, and human-robot interactions~\citep{yuan2017development}.

\section{Related Work}
\label{sec:related_works}
\subsection{Text-based Embodied AI}
Most recent works in dialogues only study the statistical regularities of language data, without an explicit understanding of the underlying world. Virtual embodiment~\citep{krishnaswamy2019multimodal} was proposed as a strategy for language research by several previous works~\citep{brooks1991intelligence, kiela2016virtual, gauthier2016paradigm, mikolov2016roadmap, lake2017building}. It implies that the best way to acquire human knowledge is to have the agent learn through experience in a situated environment. \citet{urbanek-etal-2019-learning} introduce LIGHT as a research platform for studying grounded dialogue~\citep{grice1981presupposition, grice1989indicative, stalnaker2002common}, where agents can perceive, emote, and act when conducting dialogues with other agents. \citet{ammanabrolu2020motivate} extend LIGHT with a dataset of "quests", aiming to create agents that both act and communicate with other agents in pursuit of a goal. Instead of guiding the agent to complete an in-game goal, our work aims to teach agents to speak/act in a socially intelligent way. Besides LIGHT, there are also other text-adventure game frameworks, such as~\citet{narasimhan2015language} and TextWorld~\citep{cote2018textworld}, but no human dialogues are incorporated in them. Based on the TextWorld, there are recent works~\citep{yuan2018counting, yin2019comprehensible, adolphs2019ledeepchef, adhikari2020learning} on building agents trained with reinforcement learning. 

\subsection{Mental State Transition}
An important hypothesis in the ECL~\citep{lakoff1980metaphorical, feldman2004embodied} is that humans understand the meaning of language by mentally simulating its content. Great efforts have been made to model human mental states. For example, \citet{dinan2018wizard} design a memory network capable of storing knowledge and generating natural responses conditioning on retrieved entries. \citet{adhikari2020learning} propose a graph-aided transformer agent (GATA) that infers and updates latent belief graphs during planning to enable effective action selection. However, GATA is designed for capturing game dynamics not dialogues, and our method is more flexible to encode both explicit environmental changes caused by agents' actions and implicit mental state updates triggered by agents' utterances. Such hybrid approaches mixing fixed symbolic states with deep continuous states are studied in recent neural-symbolic research~\citep{sun1994integrating, garcez2008neural, besold2017neural, yi2018neural}. The result interpretable graphs have two benefits: \textit{(i)} the mental state parsing could be viewed as a form of executable semantic parses~\citep{liang2016learning}, so it is easy to write programs to simulate the mind transition. A real-world application leveraging similar approaches is seen in~\citet{andreas-etal-2020-task}. \textit{(ii)} the unified graphical representation can be extended to model higher-order mental states, \textit{i.e.,} theory-of-mind (ToM)~\citep{premack1978does}. ToM is defined as the ability to impute mental states to oneself and others. It enables humans to make inferences about what other people believe in a given situation and predict what they will do~\citep{apperly2010mindreaders, gordon2017formal, akula2019x}. ToM is thus impossible without the capacity to form "second-order representations"~\citep{dennett1978beliefs, pylyshyn1978attribution, ganaie2015study}. 

\subsection{Human Value}
\begin{figure}[ht]
\begin{center}
\centerline{\includegraphics[width=0.8\linewidth]{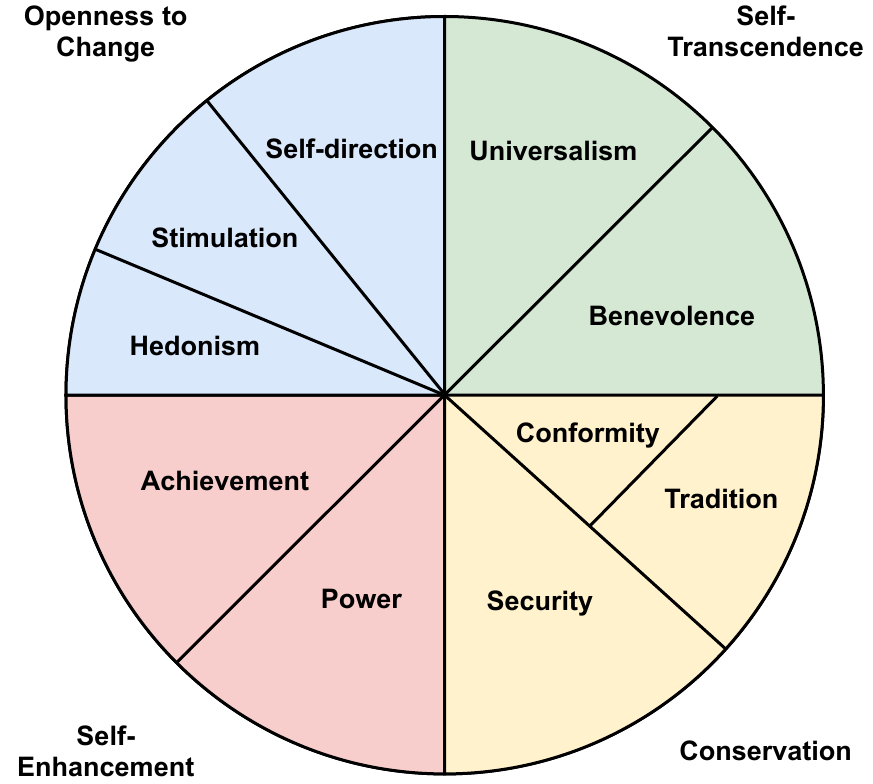}}
\caption{Theory of Basic Human Values~\citep{schwartz1992universals}.}
\label{fig:values}
\end{center}
\vspace{-8mm}
\end{figure}

When teaching agents to speak and act in a socially intelligent way, an approach considering values should be adopted. The theory of basic human values, developed by \citet{schwartz1992universals, schwartz2012overview}, tries to measure universal values that are recognized throughout major cultures. A set of $10$ basic values\footnote{A refinement of the theory~\citep{schwartz2012refining}, partitions the same continuum into 19 more narrowly defined values that permit more precise explanation and prediction.} are identified and serve as the guiding principles in the life of a person or group~\citep{ cieciuch2012comparison}, as shown in Figure \ref{fig:values}. 
Similarly, in economics and ethics, the concept of utility was developed as a measure of pleasure or satisfaction that drives human activities at all levels. Derived from the rational choice theory~\citep{abella2009soldiers}, utilitarianism states that human decision-making could be viewed as a two-step procedure. First, we select a feasible region based on financial, legal, physical, or emotional restrictions we are facing. Then we make a choice based on the preference order~\citep{allingham2002choice, de2012rational}. In this paper, we learn a transformer-based utility function of human values from the knowledge base \textsc{ValueNet}~\cite{Qiu_Zhao_Li_Lu_Peng_Gao_Zhu_2022}. Inspired by descriptive ethics, \textsc{ValueNet} provides social scenarios and annotated human preference to teach the agent human attitudes to various ethical situations. The dataset is curated from the widely used social commonsense dataset \textsc{Social-Chem-101}~\citep{forbes-etal-2020-social} and labeled with Amazon Mechanical Turk.

\section{Problem Formulation}
\label{sec:problem_form}
We will first briefly introduce the text-adventure environment LIGHT, followed by the mental state modeling and value utility formulation. 

\begin{figure*}[ht]
\begin{center}
\centerline{\includegraphics[width=\linewidth]{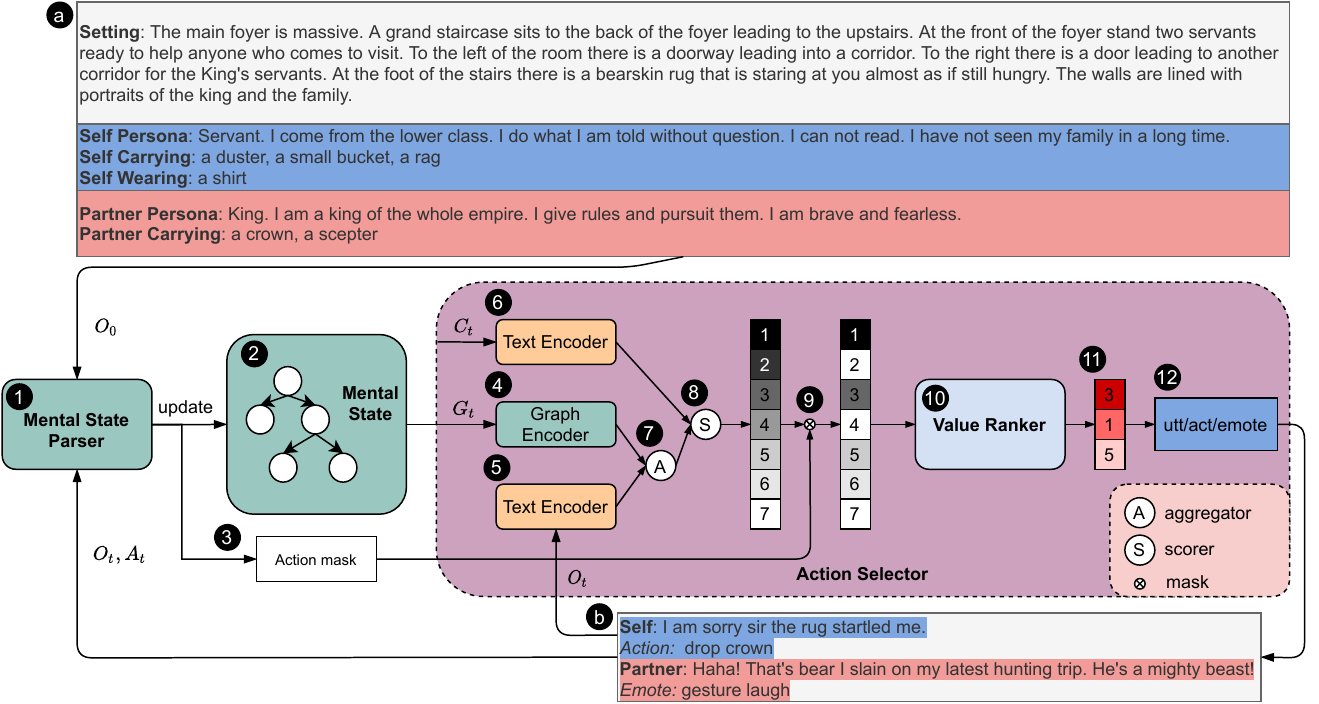}}
\caption{Socially Intelligent Agent Architecture with Mental State Parser and Value Model. }
\label{fig:mind_utility}
\end{center}
\vspace{-8mm}
\end{figure*}

\textbf{LIGHT}~\citep{urbanek-etal-2019-learning} is a large-scale crowd-sourced fantasy text-adventure platform for studying grounded dialogues. Figure \ref{fig:mind_utility}\textcircled{a} shows a typical local environment setting, including location description, objects (and their affordances), characters, and their personas. Agents can talk to other agents in free-form text, take actions defined by templates, or express certain emotions (Figure \ref{fig:mind_utility}\textcircled{b}). 
Given the environmental setting and observation history, our task is to predict the agent's utterance/action/emotion for the next turn. To achieve this goal in a socially intelligent manner, we model the agent's mental state transition and incorporate human values. The mind model is proposed to depict the agent's belief about the underlying states of the text world. Meanwhile, a utility function of human values is designed to describe human preferences in common social situations. We experiment on the text-adventure game for simplicity, but the proposed architecture supports richer environments.

\begin{figure}[th]
\begin{center}
\centerline{\includegraphics[width=\linewidth]{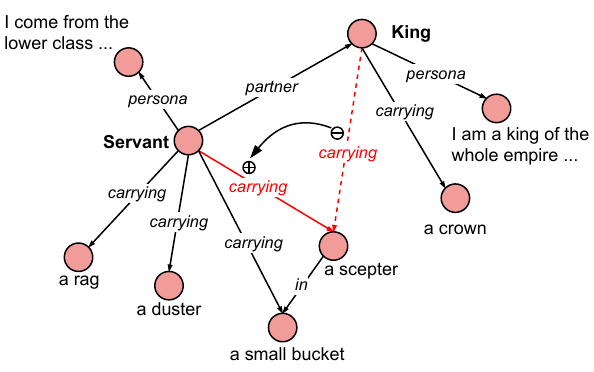}}
\caption{A graphical representation of the agent's mental state. Nodes are attributed with encoded natural language description of agents, objects and the environment. Agents' action trigger explicit topology changes of the graph.}
\label{fig:persona}
\end{center}
\vspace{-8mm}
\end{figure}
\subsection{Mental State Modeling}
Our goal is to parse, construct and maintain the mental states in dialogues. With the mental state grounding on the details of the local environment, the agent could simulate and reason the evolutionary status of the world and condition its speaking and actions. A graphical representation of the mental state is proposed, as illustrated in Figure \ref{fig:persona}. Nodes in the graph represent the involved agents, persona descriptions, objects, objects' descriptions, and setting descriptions, which will change as the game setting switches. The relational edges between these nodes describe the state of mind. The mental state is updated with the observed dialogue history or actions, \textit{e.g.}, \textit{King gives the scepter to the servant} will result in the scepter being moved from the king to the servant. 
\subsection{Human Value Modeling}
We assume that the agent in the fantasy world would make near-optimal choices to maximize the utility of its preferred values. We denote the available alternatives to be a set of $n$ exhaustive and exclusive utterances or actions $A=\{a_1,...,a_i,...,a_n\}$. The value function $f_v(\cdot)$ describes the utility score of the alternative from the value dimension $v, v\in V=\{$\textit{achievement, power, security, conformity, tradition, benevolence, universalism, self-direction, stimulation, hedonism}$\}$\footnote{Detailed definition for each dimension is attached in Appendix~\ref{sec:a.1}.}. For example, if $a_i$ is more preferred than $a_j$ in terms of \textit{security}, then $f_{security}(a_i)>f_{security}(a_j).$ Usually, we cannot find an analytical form of the value function. However, what matters for preference ordering is which of the two options gives the higher expected utility, not the numerical values of those expected utilities.

In LIGHT, the agent's value priority is reflected by its persona description. For the example in Figure \ref{fig:mind_utility}\textcircled{a}, the servant is a person who values \textit{conformity} and \textit{tradition} and has a lower priority on \textit{self-direction} and \textit{stimulation}. Using the same value function to approximate a value priority parser: $f_v(p)$, where $p$ is the persona description, the utility or the desirability of candidate $a_i$ to person $p$ is the Euclidean distance between its value priority and the candidate's utility score:
\begin{equation}
u(a_i) = \sqrt {\sum _{v\in V} \left( f_v(p)-f_v(a_i)\right)^2 }.
\label{eq:utility}
\end{equation}
Since some actions could be impossible physically (\textit{e.g., one cannot drop an object if the agent is not carrying the object}), the decision making process becomes a problem of maximizing the utility score that is subject to some constraints from the mental state, \textit{i.e.}, $u(a|c)$, where $c$ represents the context or constraints. 

\section{Algorithms}
\label{sec:alg}
\begin{figure*}[ht]
\begin{center}
\centerline{\includegraphics[width=0.9\linewidth]{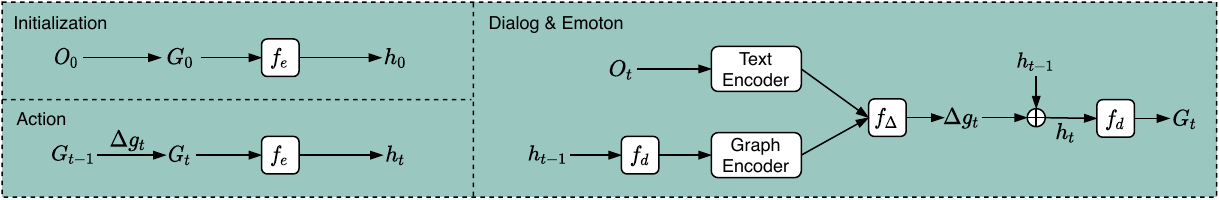}}
\caption{Overall Architecture of the Hybrid Mental State Parser}
\label{fig:parser}
\end{center}
\vspace{-8mm}
\end{figure*}
The overall architecture of our proposed framework is illustrated in Figure \ref{fig:mind_utility}. For each scenario, a setting description (Figure \ref{fig:mind_utility}\textcircled{a}) is provided by the LIGHT environment, which can include a description of the location, object affordances, agents' personas, and the objects that agents are carrying, wearing, or wielding. The free-form conversations, actions, and emotions are logged during the communication as the observation history (Figure \ref{fig:mind_utility}\textcircled{b}). To begin with, a mental state parser will parse the setting descriptions into graph representation and initialize the agent's mental state (steps \textcircled{\scriptsize{1}} and \textcircled{\scriptsize{2}}). Besides the mental state updating, the parser also outputs an action mask that is aimed to rule out actions that are physically or causally impossible to take (step \textcircled{\scriptsize{3}}). A graph encoder (step \textcircled{\scriptsize{4}}) and a text encoder (step \textcircled{\scriptsize{5}}) will convert the mental state graph $G_t$ and the dialogue observation $O_t$ into vector representations, respectively. The same text encoder will be used to encode the candidates $C_t$ (step \textcircled{\scriptsize{6}}). In step \textcircled{\scriptsize{7}}, the context vectors are combined by a bi-directional attention aggregator~\citep{yu2018qanet, seo2016bidirectional}, and each candidate is assigned a score with a Multi-Layer Perceptron (MLP) (step \textcircled{\scriptsize{8}}). The action mask is then applied to get the feasible candidates under current mental state constraints (step \textcircled{\scriptsize{9}}). In steps \textcircled{\scriptsize{10}} and \textcircled{\scriptsize{11}}, the top three candidates from the last step will be fed into the value model and re-ranked. Finally, the selected utterance/action/emotion is executed by the agent (step \textcircled{\scriptsize{12}}) and fed back to the environment. Upon receiving the response from other agents in the environment, the new observation will be again parsed and used to update the agent's state of mind, and the cycle repeats. In the following, we will describe each component in more detail.

\subsection{Mental State Modeling (steps \textcircled{\scriptsize{1}}-\textcircled{\scriptsize{2}})}
Figure \ref{fig:parser} describes the architecture of the mental state parser. 
We define the mental state graph  $G\in [-1,1]^{R\times N \times N}$, where $R$ is the maximum number of relation types and $N$ is the maximum number of entities. 
The initial mental state graph $G_0$ is constructed by a ruled-based parser from the setting description $O_0$. The graph is encoded by function $f_e$ to a hidden state $h_0$ that is later used for graph update. At game step $t$, the mental state parser parses relevant information from observation $O_t$ and update the agent's mental state from $G_{t-1}$ to $G_t$. Considering that observation $O_t$ typically conveys incremental information from step $t-1$ to $t$, we generate the graph update $\Delta g_t$ instead of the whole graph at each step
\begin{equation}
    G_t = G_{t-1} \oplus \Delta g_t, 
\end{equation}
where $\oplus$ is the graph update operation. 
The graph update can be either discrete or continuous, and there have been studies on the pros and cons of each updating method~\citep{adhikari2020learning}. The discrete approach may suffer from an accumulation of errors but benefit from its interpretability. The continuous graph model needs to be trained from data, but it is more robust to possible errors. In this work, we propose a hybrid (discrete-continuous) method for updating the agent's state of mind by considering there exists a mixture of discrete events and continuous information in typical human-machine interactive environments. In the specific example of our tested LIGHT, the actions or events are template-based, it is more appropriate to adopt a discrete method for parsing; meanwhile, since utterances are challenging to be encoded into discrete representations, we apply a continuous update method instead.

\subsubsection{Discrete Graph Definition \& Update}
To update the graph, we define $\Delta g_t$ as a sequence of update operations of the following two atomic types:
\begin{itemize}
    \item ADD(\texttt{src}, \textcolor{black}{\texttt{dst}}, \textcolor{black}{\texttt{relation}}): add a directed edge, named \textcolor{black}{\texttt{relation}}, from node \textcolor{black}{\texttt{src}} to node \textcolor{black}{\texttt{dst}}.
    \item DEL(\textcolor{black}{\texttt{src}}, \textcolor{black}{\texttt{dst}}, \textcolor{black}{\texttt{relation}}): delete a directed edge, named \textcolor{black}{\texttt{relation}}, from node \textcolor{black}{\texttt{src}} to node \textcolor{black}{\texttt{dst}}.
\end{itemize}
LIGHT defines various actions including \textit{get}, \textit{drop}, \textit{put}, \textit{give}, \textit{steal}, \textit{wear}, \textit{remove}, \textit{eat}, \textit{drink}, \textit{hug} and \textit{hit}, and each taking either one or two arguments, \textit{e.g.}, \textit{give scepter to servant}. Every action could be parsed as one or a sequence of update operators that act on $G_{t-1}$. For example, actor performing ``\textit{give object to agent}'' can be parsed into DEL(\textit{actor}, \textit{object}, \textit{carrying}) and ADD(\textit{agent}, \textit{object}, \textit{carrying}). 
The rule-based parsing of the setting description and the discrete events could also be replaced by a seq2seq decoding process. Since both strings are well-structured in LIGHT, we omit training such a decoder for simplicity. Note that actions in LIGHT could only be executed when constraints are met, so we also generate an action mask according to the current mental state. By checking the adjacency matrix, we rule out action candidates conducted on objects that are inaccessible.

\subsubsection{Continuous Graph Definition \& Update}
Besides the actions taken by the agents, their utterances could also have an implicit impact on the agents' mental states. To handle the continuous dialogue observation, we use a recurrent neural network as the graph update operation $\oplus$. 
\begin{equation}
    \begin{aligned}
        \Delta g_t &= f_{\Delta}(h_{G_{t-1}},h_{O_t}), \\
        h_t &= \text{RNN}(\Delta g_t, h_{t-1}),  \\
        G_t &= \text{MLP}(h_t).
    \end{aligned}
\end{equation}
The function $f_{\Delta}$ aggregates the information from the previous mental state $G_{t-1}$ and observation $O_t$ to generate the graph update $\Delta g_t$. $h_{G_{t-1}}$ denotes the representation of $G_{t-1}$ from the graph encoder. $h_{O_t}$ is the output of the text encoder. $h_t$ is a hidden state acting as the memory, from which we decode the new mental state $G_t$ using a MLP. For the recurrent operator, we could either use LSTM~\citep{hochreiter1997long} or GRU~\citep{cho2014properties}. 
More details on the graph encoder and text encoder we applied are presented in the section \ref{subsec:language&action}.

\subsection{Action Selector (steps \textcircled{\scriptsize{4}}-\textcircled{\scriptsize{11}})}
\label{subsec:language&action}
Conditioned on the agent's mental state, the action selector chooses the optimal candidate based on the prediction task (\textit{i.e.}, utterance, action, or emotion). The selector consists of five components: a graph encoder (Fig. \ref{fig:mind_utility}\textcircled{\scriptsize{4}}) to convert the state-of-mind graph to a hidden state vector; a text encoder (Fig. \ref{fig:mind_utility}(\textcircled{\scriptsize{5}}, \textcircled{\scriptsize{6}})) to encode the dialogue history and text candidates; an aggregator (Fig. \ref{fig:mind_utility}\textcircled{\scriptsize{7}}) to fuse the two context representations; a general scorer (Fig. \ref{fig:mind_utility}\textcircled{\scriptsize{8}}) to assign a score to each candidate; and a value model (Fig. \ref{fig:mind_utility}\textcircled{\scriptsize{10                            }}) to re-rank the candidates based on the assigned persona.

\textbf{1. Graph Encoder.} We use relational graph convolutional networks (R-GCNs)~\citep{schlichtkrull2018modeling} to encode the graph representation of mental states. The R-GCN is adapted from Graph Convolutional Networks (GCNs) so that it could embed the edge attributes (relational text embedding) in the mental state graph.

\textbf{2. Text Encoder.} A BERT-based~\cite{devlin-etal-2019-bert} encoder converts the text-based dialogue history into a vector representation, using the last hidden state corresponding to the \texttt{[CLS]} token; We also use the same encoder to encode the text response candidates. 

\textbf{3. Aggregator.} A bi-directional attention layer~\citep{yu2018qanet, seo2016bidirectional} is adopted to fuse the information from the mental state and the contextualized text hidden state. The co-attention allows the agent to focus on the memory part that has been mentioned in the dialogue.

\textbf{4. Scorer.} The full context representation vector is concatenated with each candidate and an MLP layer with softmax activation generates a score for each of them. 
\begin{figure}[ht]
\begin{center}
\centerline{\includegraphics[width=\linewidth]{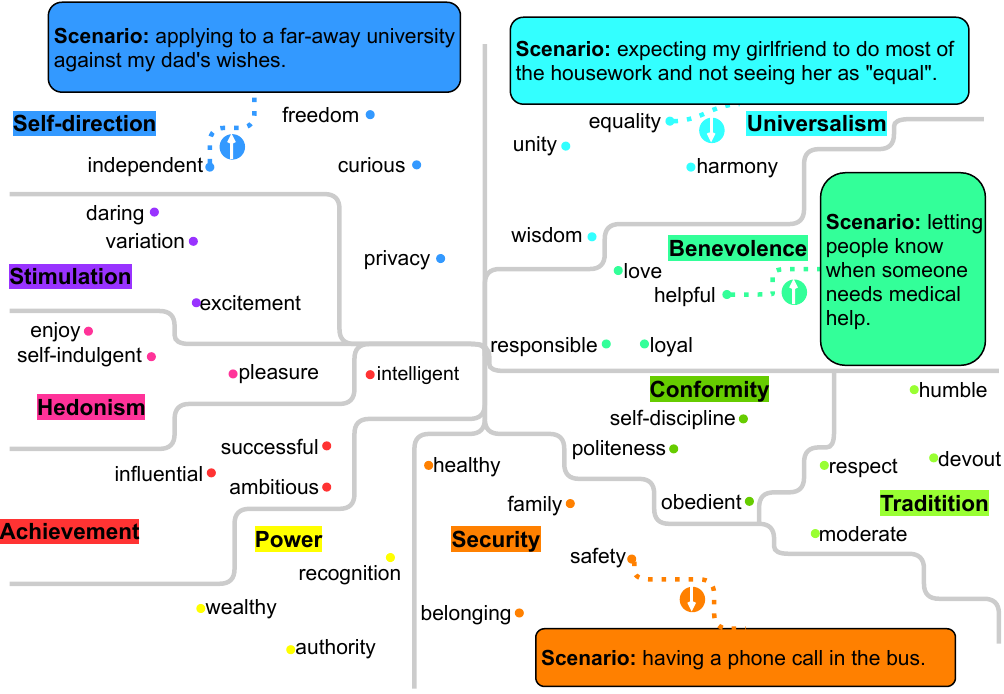}}
\caption{The \textsc{ValueNet}~\cite{Qiu_Zhao_Li_Lu_Peng_Gao_Zhu_2022} dataset with social scenarios organized by Schwartz values~\cite{schwartz2012overview}.}
\label{fig:valuenet}
\end{center}
\vspace{-8mm}
\end{figure}
\begin{table*}[]
\centering
\resizebox{0.8\linewidth}{!}{%
\begin{tabular}{l|ccc|ccc}
\toprule
                                             & \multicolumn{3}{c|}{\textit{Seen Test}}                                                         & \multicolumn{3}{c}{\textit{Unseen Test}}                                                      \\
                                             & \multicolumn{1}{c}{Dialogue} & \multicolumn{1}{c}{Action} & \multicolumn{1}{c|}{Emotion} & \multicolumn{1}{c}{Dialogue} & \multicolumn{1}{c}{Action} & \multicolumn{1}{c}{Emotion} \\
Method                                       & \multicolumn{1}{c}{R@1/20}   & \multicolumn{1}{c}{Acc}    & \multicolumn{1}{c|}{Acc}   & \multicolumn{1}{c}{R@1/20}   & \multicolumn{1}{c}{Acc}    & \multicolumn{1}{c}{Acc}   \\ \midrule
BERT-based Bi-Ranker                         & 76.5                        & 42.5                      & 25.0                      & 70.5                        & 38.8                      & 25.7                     \\
BERT-based Cross-Ranker                      & 74.9                        & 50.7                      & 25.8                      & 69.7                        & 51.8                      & 28.6                     \\
\hline
discrete mental state                                & 75.8                        & 52.1                      & 25.1                      & 69.9                        & 53.4                      & 25.5                      \\
continuous mental state                                   & 77.3                        & 49.3                      & \textbf{26.2}                      & 72.1                        & 45.2                      & 29.1                      \\
hybrid mental state                          & 78.4                        & 53.5                      & 26.1                      & 72.3                        & 54.3                      & 29.5                      \\
hybrid+mask                     & 78.5                        & 54.5                      & 26.1                      & 72.3                        & 55.4                      & 29.4                      \\
hybrid+mask+value & \textbf{78.8}                        & \textbf{56.4}                      & 26.1                      & \textbf{72.6}                        & \textbf{57.5}                      & \textbf{30.1}                      \\ \midrule
Human Performance*                           & 87.5                        & 62.0                      & 27.0                      & 91.8                        & 71.9                      & 34.4                     \\ \bottomrule
\end{tabular}%
}
\caption{Model performance on the LIGHT \textit{Seen Test} and \textit{Unseen Test}. For dialogue prediction, Recall@1/20 is reported for ranking the ground truth among 19 other randomly chosen candidates. Percentage accuracy is calculated for action and emotion prediction. (*) Human performance is reported by the original paper~\citep{urbanek-etal-2019-learning} on a subset of data.}
\label{tab:result}
\end{table*}

\textbf{5. Value Ranker.} After all the candidates are ranked, we select the top three candidates and then re-rank them according to the proposed value model. The value model is a BERT-based utility scorer trained on the knowledge base \textsc{ValueNet}~\cite{Qiu_Zhao_Li_Lu_Peng_Gao_Zhu_2022}.
A custom input format constructed as `\texttt{[CLS][\$VALUE]s}' is fed into the BERT, \textit{i.e.,}
\begin{equation}
    f_{v}(\texttt{s}) = \textsc{BERT}(\texttt{[CLS][\$VALUE]s}),  
\end{equation}
where \texttt{[CLS]} is the special token for regression, $\texttt{s}$ is the scenario, and \texttt{[\$VALUE]} are special tokens we define to prompt~\cite{li2021prefix, brown2020language} the transformer the interested value dimension $v$. A regression head is put on top of the model to get a continuous estimation of the utility in the range of $[-1, 1]$.

The \textsc{ValueNet} is organized in $10$ dimensions of Schwartz values, as shown in Figure \ref{fig:valuenet}. It consists of social scenarios curated from \textsc{Social-Chem-101}~\citep{ forbes-etal-2020-social}. And the samples are annotated by Amazon Mechanical Turk workers, who are asked about their attitudes towards provided scenarios. For example, if you are someone who values \textit{benevolence}, will you do or say: ``today I buried and mourned a rat"? Their choices (yes, no, unrelated) are then quantified to numerical utilities: +1, -1, 0, respectively. 

\section{Experiments}
\label{sec:exp}
We conduct experiments on the LIGHT dataset and compare our model with state-of-the-art methods based on two variants of BERT models. An ablation study is carried out to justify our model design, and a case study is performed to demonstrate how the proposed framework could help the agent ground upon the environment details and make value-driven decisions. 

\subsection{Experimental Setup and Implementation} 
The dialogues in LIGHT are split into \textit{train} (8539), \textit{valid} (500), \textit{seen test} (1000), and \textit{unseen test} (739) as the dataset is released. The \textit{unseen test} set consists of dialogues collected on a set of scenarios that have not appeared in the training data.
We use the history of dialogues, actions, and emotions to predict the agent's next turn. Note that the original paper manually filters out actions with no affordance leveraging the object annotation, while we provide all candidates to demonstrate our model's capability of reasoning feasible actions automatically from the agent's mental state.

\begin{figure}[ht]
\begin{center}
\centerline{\includegraphics[width=\linewidth]{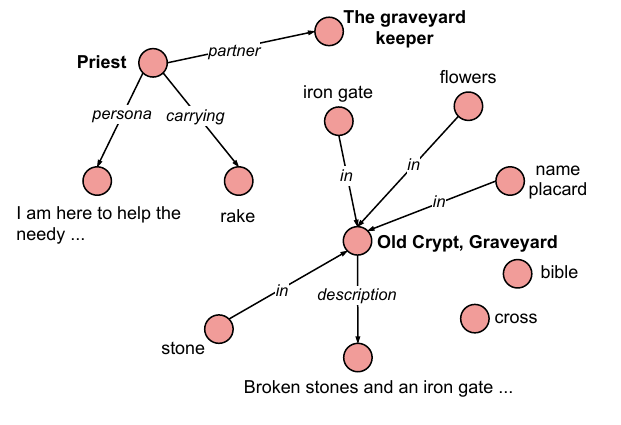}}
\caption{Initial mental state graph parsed from the example setting string in Appendix~\ref{sec:a.1}. The nodes of objects' descriptions are omitted to save space.}
\label{fig:init_graph}
\end{center}
\vspace{-8mm}
\end{figure}

Here we describe the implementation details of the proposed framework. The mental state graph is initialized with a structured setting string including all involved elements in the scenario (an example is attached in Appendix~\ref{sec:a.2}). The setting parser is based on general parsing tools: regular expression and spaCy~\citep{spacy2, clark-manning-2016-deep, honnibal-johnson-2015-improved}, resulting in the initial mental state graph as shown in Figure \ref{fig:init_graph}. 
For the functions $f_e$ and \textbf{$f_d$}, we use two-layer MLPs with tanh~\citep{karlik2011performance} and ReLU~\citep{agarap2018deep} activations. The \textbf{Text Encoder} is a pretrained BERT (base-uncased) model~\citep{wolf-etal-2020-transformers}. The \textbf{Graph Encoder} is an R-GCN with six layers and a hidden size of 64. We also adopt the highway connections between consecutive layers for faster convergence and 3-basis decomposition to reduce the parameters and prevent overfitting. 

\subsection{Baseline Models} 
Two BERT-based models~\cite{urbanek-etal-2019-learning} are used as strong baselines, which have kept the state-of-the-art performance on this task. \textbf{BERT Bi-Ranker} produces a vector representation for the context and each candidate. 
Each candidate is assigned a score by the dot product between the context embedding and the candidate embedding. 
\textbf{BERT Cross-Ranker} concatenates the context string with each candidate and feeds the string to the BERT model instead. 
Compared with the bi-ranker, The cross-ranker allows the model to attend to the context when encoding each candidate.

\subsection{Results and Analysis}
Table \ref{tab:result} shows the results, where our model outperforms the state-of-the-art models by a large margin. To understand the results, we first compare mental state graph designs using discrete, continuous, and the proposed hybrid parser. 

The discrete mental state parser uses actions to explicitly update the graph to augment the context representation. In the action prediction task, the discrete parser outperforms the purely continuous method (+2.8\% (seen), +8.2\% (unseen)), the BERT Bi-Ranker (+9.6\% (seen), +14.6\% (unseen)), and the BERT Cross-Ranker (+1.4\% (seen), +1.6\% (unseen)). While the continuous mental state parser misses the hard constraints introduced by less frequent actions, it updates the graph implicitly with the dialogues and shows a better result than the discrete one on dialogue prediction (+1.5\% (seen), +2.2\% (unseen)) and emotion prediction (+1.1\% (seen), +3.6\% (unseen)). 

The hybrid mental state parser performs the best among the three according to almost all metrics, mainly because it aggregates the soft update from the dense dialogue and the hard constraints from the sparse actions. We also notice that the emotion prediction in LIGHT is a hard task because it is not strictly constrained by the context. Even humans can only achieve 27.0\% (seen) and 34.4\% (unseen) accuracy. Nevertheless, our model provides a relatively 1.2\% (seen) and 3.1\% (unseen) performance boost compared to the best BERT baseline.

\begin{figure}[ht]
\begin{center}
\centerline{\includegraphics[width=\linewidth]{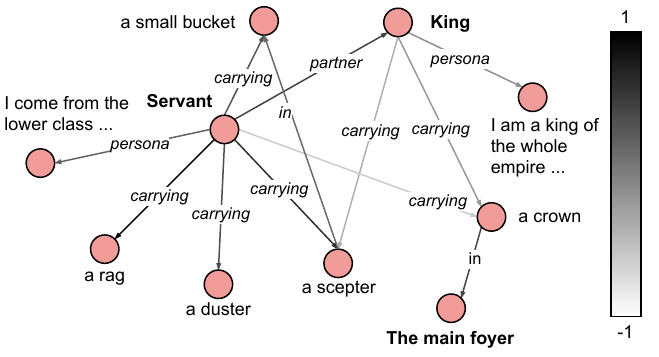}}
\caption{Intermediate mental state for the agent \textbf{Servant} in the dialogue example of Figure \ref{fig:mind_utility}. The adjacency matrix of the mental state graph is visualized and the darkness of the edges represent the relation strength. Only critical relation types between nodes are shown for illustration purpose.}
\label{fig:inter_graph}
\end{center}
\vspace{-4mm}
\end{figure}
Then, with the ablation study of our proposed action mask (hybrid mental state \textit{vs.} hybrid+mask), we prove the effectiveness of it for improving action accuracy by $\sim$1\% in action prediction. Figure \ref{fig:inter_graph} demonstrates how the mental state could help agent ground on the context. We can see a very weak relation of the type "\textit{carrying}" between the agent servant and the object crown. Thus the servant should not be able to give the crown to others at this time step. Though our model does not rely on annotated action affordances during action predicting, an action mask can be reasoned from such a mental state, which helps filter out physical or causally impossible actions. 

Lastly, we analyze the results after introducing the value model. 
We first compute the value priority of the agent by applying the value function to its persona description. For example, given the servant's persona description in Figure~\ref{fig:mind_utility}, it shows \textit{conformity, tradition}, and \textit{security} have higher utility scores to the agent than other dimensions. Then we calculate utility scores of the top three candidates based on Equation~\ref{eq:utility}. This teaches the agent to make decisions that align with the assigned role and further improves the overall performance, (+0.3\% (seen), +0.3\% (unseen)) for dialogue prediction, (+1.9\% (seen), +2.1\% (unseen)) for action prediction, and +0.7\% (unseen) for emotion prediction.

\section{Conclusion}
\label{sec:conclusion}
This paper proposes to build a socially intelligent agent by incorporating mind simulation and human values. We explore using a hybrid parser to track agents' mental state transition. The value model pretrained on \textsc{ValueNet} brings social preference to help the agent make decisions. The model is proved to have a better performance than the state-of-the-art models on LIGHT. In the future, we have a plan to build a dataset to study the implicature in conversation and model deeper levels in the Theory of Mind based on the proposed representation.

\bibliography{anthology,custom}
\bibliographystyle{acl_natbib}

\appendix

\section{Appendix}
\label{sec:appendix}
\subsection{Schwartz Value Definition}
\label{sec:a.1}
\textbf{Self-Direction}
Defining goal: independent thought and action--choosing, creating, exploring. Self-direction derives from organismic needs for control and mastery and interactional requirements of autonomy and independence. (creativity, freedom, choosing own goals, curious, independent) [self-respect, intelligent, privacy]

\textbf{Stimulation}
Defining goal: excitement, novelty, and challenge in life. Stimulation values derive from the organismic need for variety and stimulation in order to maintain an optimal, positive, rather than threatening, level of activation. This need probably relates to the needs underlying self-direction values. (a varied life, an exciting life, daring)

\textbf{Hedonism}
Defining goal: pleasure or sensuous gratification for oneself. Hedonism values derive from organismic needs and the pleasure associated with satisfying them. Theorists from many disciplines mention hedonism. (pleasure, enjoying life, self-indulgent)

\textbf{Achievement}
Defining goal: personal success through demonstrating competence according to social standards. Competent performance that generates resources is necessary for individuals to survive and for groups and institutions to reach their objectives. As defined here, achievement values emphasize demonstrating competence in terms of prevailing cultural standards, thereby obtaining social approval. (ambitious, successful, capable, influential) [intelligent, self-respect, social recognition]

\textbf{Power}
Defining goal: social status and prestige, control or dominance over people and resources. The functioning of social institutions apparently requires some degree of status differentiation. A dominance/submission dimension emerges in most empirical analyses of interpersonal relations both within and across cultures. To justify this fact of social life and to motivate group members to accept it, groups must treat power as a value. Power values may also be transformations of individual needs for dominance and control. Value analysts have mentioned power values as well. (authority, wealth, social power) [preserving my public image, social recognition]

Both power and achievement values focus on social esteem. However, achievement values (e.g., ambitious) emphasize the active demonstration of successful performance in concrete interaction, whereas power values (e.g., authority, wealth) emphasize the attainment or preservation of a dominant position within the more general social system.

\textbf{Security}
Defining goal: safety, harmony, and stability of society, of relationships, and of self. Security values derive from basic individual and group requirements. Some security values serve primarily individual interests (e.g., clean), others wider group interests (e.g., national security). Even the latter, however, express, to a significant degree, the goal of security for self or those with whom one identifies. (social order, family security, national security, clean, reciprocation of favors) [healthy, moderate, sense of belonging]

\textbf{Conformity}
Defining goal: restraint of actions, inclinations, and impulses likely to upset or harm others and violate social expectations or norms. Conformity values derive from the requirement that individuals inhibit inclinations that might disrupt and undermine smooth interaction and group functioning. As I define them, conformity values emphasize self-restraint in everyday interaction, usually with close others. (obedient, self-discipline, politeness, honoring parents and elders) [loyal, responsible]

\textbf{Tradition}
Defining goal: respect, commitment, and acceptance of the customs and ideas that one's culture or religion provides. Groups everywhere develop practices, symbols, ideas, and beliefs that represent their shared experience and fate. These become sanctioned as valued group customs and traditions. They symbolize the group's solidarity, express its unique worth, and contribute to its survival (Durkheim, 1912/1954; Parsons, 1951). They often take the form of religious rites, beliefs, and norms of behavior. (respect for tradition, humble, devout, accepting my portion in life) [moderate, spiritual life]

Tradition and conformity values are especially close motivationally; they share the goal of subordinating the self to socially imposed expectations. They differ primarily in the objects to which one subordinates the self. Conformity entails subordination to persons with whom one frequently interacts—parents, teachers, and bosses. Tradition entails subordination to more abstract objects—religious and cultural customs and ideas. As a corollary, conformity values exhort responsiveness to current, possibly changing expectations. Tradition values demand responsiveness to immutable expectations from the past.

\textbf{Benevolence}
Defining goal: preserving and enhancing the welfare of those with whom one is in frequent personal contact (the ‘in-group’). Benevolence values derive from the basic requirement for smooth group functioning and from the organismic need for affiliation. Most critical are relations within the family and other primary groups. Benevolence values emphasize voluntary concern for others’ welfare. (helpful, honest, forgiving, responsible, loyal, true friendship, mature love) [sense of belonging, meaning in life, a spiritual life].

Benevolence and conformity values both promote cooperative and supportive social relations. However, benevolence values provide an internalized motivational base for such behavior. In contrast, conformity values promote cooperation in order to avoid negative outcomes for self. Both values may motivate the same helpful act, separately or together.

\textbf{Universalism}
Defining goal: understanding, appreciation, tolerance, and protection for the welfare of all people and for nature. This contrasts with the in-group focus of benevolence values. Universalism values derive from survival needs of individuals and groups. But people do not recognize these needs until they encounter others beyond the extended primary group and until they become aware of the scarcity of natural resources. People may then realize that failure to accept others who are different and treat them justly will lead to life-threatening strife. They may also realize that failure to protect the natural environment will lead to the destruction of the resources on which life depends. Universalism combines two subtypes of concern—for the welfare of those in the larger society and world and for nature (broadminded, social justice, equality, world at peace, world of beauty, unity with nature, wisdom, protecting the environment)[inner harmony, a spiritual life]

\subsection{Example Environment Setting}
\label{sec:a.2}
An example setting string for the utterance prediction is:\\
"\textbf{\_task\_speech} \\
\textbf{\_setting\_name} Old Crypt, Graveyard \\
\textbf{\_setting\_desc} Broken stones and a iron gate closing the entrance with a name placard that the name is worn off. \\
\textbf{\_partner\_name} the graveyard keeper who lives across the yard
\textbf{\_self\_name} priest\\
\textbf{\_self\_persona} I am here to help the needy. I am well respected in the town. I can not accept lying.\\
\textbf{\_object\_desc} a gate : The gate is made out of rusty metal. It squeaks as it swings on its hinges.\\
\textbf{\_object\_desc} a flowers : you can see them up close but not afar. when noticed, you realize that they are old.\\
\textbf{\_object\_desc} a name placard : The placard is made of wood witha clear name on it.\\
\textbf{\_object\_desc} a stone : The stone is chipped from being used as target practice from soldier trainees\\
\textbf{\_object\_desc} a placard : A sign used to display names of buildings or notices.\\
\textbf{\_object\_desc} an iron gate : The gate is ornate, with complicated iron scrollwork patterns.\\
\textbf{\_object\_desc} a Rake : This rake is made of carefully split wood with a sturdy looking handle. Seems useful for keeping the leaves under control.\\
\textbf{\_object\_desc} a Cross : The cross is broken and with a few dents in the sides.\\
\textbf{\_object\_desc} a bible : The bible is bound by black leather, its pages yellowed by years of use.\\
\textbf{\_object\_in\_room} a gate\\
\textbf{\_object\_in\_room} a flowers\\
\textbf{\_object\_in\_room} a name placard\\
\textbf{\_object\_in\_room} a stone\\
\textbf{\_object\_in\_room} a placard\\
\textbf{\_object\_in\_room} an iron gate\\
\textbf{\_object\_carrying} a Rake".\\
The result mental state graph parsed from this setting is illustrated in Figure~\ref{fig:init_graph}.

\end{document}